\documentclass[final]{cvpr}

\usepackage{times}
\usepackage{epsfig}
\usepackage{graphicx}
\usepackage{amsmath}
\usepackage{amssymb}

\usepackage{subcaption}
\usepackage{multirow}
\usepackage{booktabs}
\usepackage{array}
\newcolumntype{P}[1]{>{\centering\arraybackslash}p{#1}}
\usepackage{makecell}

\usepackage{caption}
\usepackage{xcolor}

\definecolor{colormark}{RGB}{0,0,255}

\usepackage[pagebackref=true,breaklinks=true,colorlinks,bookmarks=false]{hyperref}

\newcommand\norm[1]{\left\lVert#1\right\rVert}

\def\cvprPaperID{5052} 
\def\confYear{CVPR 2021}
\begin{document}

\title{Capsule Network is Not More Robust than Convolutional Network}

\author{Jindong Gu$^1$, Volker Tresp$^1$, Han Hu$^2$\\
University of Munich$^1$\\
Microsoft Research Asia$^2$\\
{\tt\small jindong.gu@outlook.com, volker.tresp@siemens.com, hanhu@microsoft.com}
}

\maketitle

\begin{abstract}
The Capsule Network is widely believed to be more robust than Convolutional Networks. However, there are no comprehensive comparisons between these two networks, and it is also unknown which components in the CapsNet affect its robustness. In this paper, we first carefully examine the special designs in CapsNet that differ from that of a ConvNet commonly used for image classification. The examination reveals five major new/different components in CapsNet: a transformation process, a dynamic routing layer, a squashing function, a marginal loss other than cross-entropy loss, and an additional class-conditional reconstruction loss for regularization. Along with these major differences, we conduct comprehensive ablation studies on three kinds of robustness, including affine transformation, overlapping digits, and semantic representation. The study reveals that some designs, which are thought critical to CapsNet, actually can harm its robustness, i.e., the dynamic routing layer and the transformation process, while others are beneficial for the robustness. Based on these findings, we propose enhanced ConvNets simply by introducing the essential components behind the CapsNet's success. The proposed simple ConvNets can achieve better robustness than the CapsNet.
\end{abstract}

\section{Introduction}
The Capsule network (CapsNet)~\cite{sabour2017dynamic} was proposed to address the intrinsic limitations of convolutional networks (ConvNet)~\cite{lecun1998gradient}, such as the exponential inefficiency and the lack of robustness to affine transformations. In recent years, It has been sugested that CapsNets have the potential to surpass the dominant convolutional networks in these aspects~\cite{sabour2017dynamic,hinton2018matrix,rajasegaran2019deepcaps,gu2020improving,de2020introducing,mazzia2021efficient}. However, there lack comprehensive comparisons to support this assumption, and even for some reported improvements, there are no solid ablation studies to figure out which ones of the components in CapsNets are, in fact, effective.

In this paper, we first carefully examine the major differences in design between the capsule networks and the common convolutional networks adopted for image classification. A common convolutional network follows a simple algorithm flow, using a backbone convolutional network to extract image features, a global average pooling layer plus a linear layer to produce the classification logits (or optionally several fully connected layers~\cite{krizhevsky2017imagenet}), and an $N$-way SoftMax loss to drive the learning. To be better aligned with the capsule (vector) representations, the capsule networks introduce several special components. These components involve (see Fig.~\ref{fig:overview} for detailed architectures):
\begin{itemize}
    \item a non-shared transformation module, in which the primary capsules are transformed to execute votes by non-shared transformation matrices; 
    \item a dynamic routing layer to automatically group input capsules to produce output capsules with high agreements in each output capsule; 
    \item a squashing function, which is applied to \textit{squash} the capsule vectors such that their lengths distribute in the range of $[0, 1)$;
    \item a marginal classification loss to work together with the \emph{squashed} capsule representations;
    \item a class-conditional reconstruction sub-network with a reconstruction loss, targeting at recovering the original image from the capsule representations. This sub-network acts as a regularization force, in complementary to the classification loss.
\end{itemize}

Unlike previous studies~\cite{sabour2017dynamic,hinton2018matrix} which usually takes CapsNet as a whole to test its robustness, we instead try to study the effects of each of the above components in their effectiveness on robustness. We consider the three different aspects shown in~\cite{sabour2017dynamic}:
\begin{itemize}
    \item the robustness to affine transformations,
    \item the ability to recognizing overlapping digits,
    \item the semantic representation compactness.
\end{itemize}

Our investigations reveal that some widely believed benefits of Capsule networks could be wrong:
\begin{enumerate}
    \item  The ConvNets baseline adopted in comparison with CapsNets is weak~\cite{sabour2017dynamic}. Concretely, there is no global average pooling layer before the classification head in this baseline, which sacrifices the ability of spatial invariance to some extent and is harmful for generalization to novel views. In fact, a ConvNet with an additional global average pooling layer can outperform CapsNet by a large margin in the robustness to affine transformation;
    \item The dynamic routing actually may harm the robustness to input affine transformation, in contrast to the common belief;
    \item The high performance of CapsNets to recognize overlapping digits can be mainly attributed to the extra modeling capacity brought by the transformation matrices.
    \item Some components of CapsNets are indeed beneficial for learning semantic representations, e.g., the conditional reconstruction and the squashing function, but they are mainly auxiliary components and can be applied beyond CapsNets.
\end{enumerate}

In addition to these findings, we also enhance common ConvNets by the useful components of CapsNet, and achieve greater robustness. The paper is organized as follows: Sec.~\ref{sec:related_work} introduces the CapsNet and related work. In Sec.~\ref{sec:study}, we examine the behavior of CapsNets and ConvNets on three kinds of robustness, one by one, and component by component. The last section concludes our work and discusses future work.

\section{Background and Related Works}
\label{sec:related_work}
\begin{figure*}[t]
    \centering
        \includegraphics[width=1.01\textwidth]{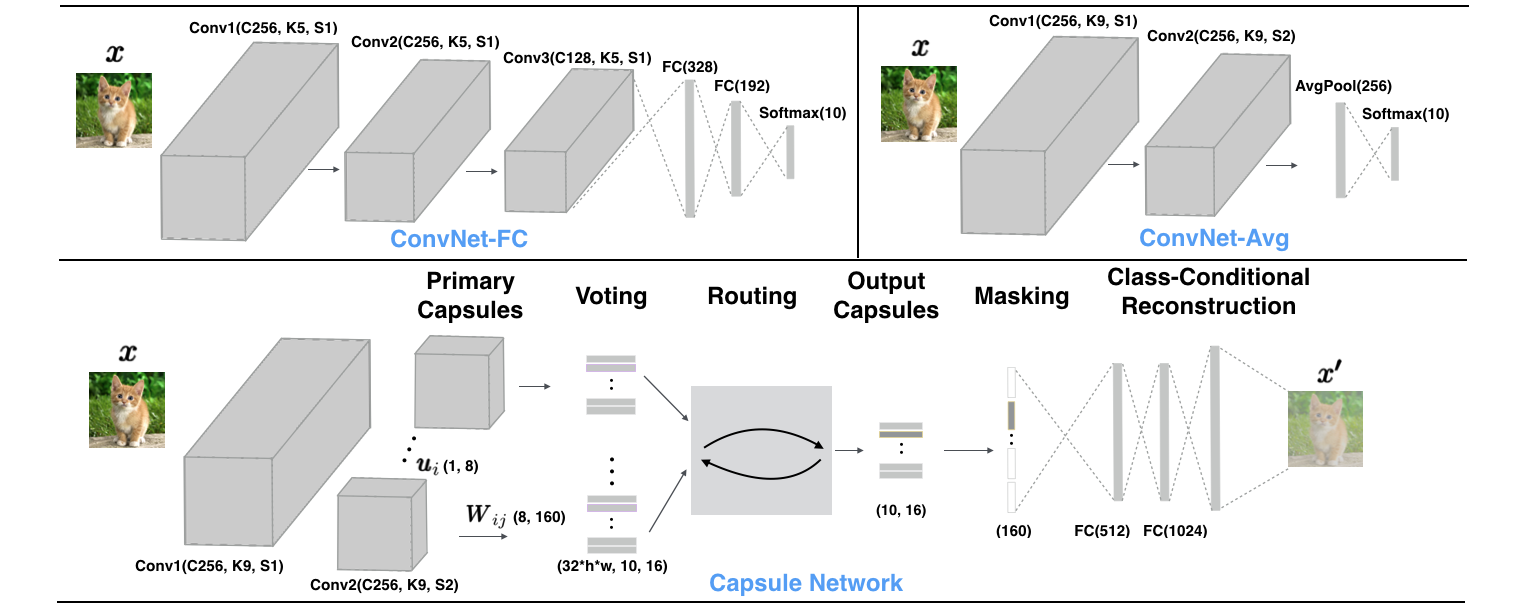}
        \caption{The overview of ConvNet and CapsNet architectures: The ConvNet-FC is a naive ConvNet architecture, while ConvNet-Avg is the one commonly used in image classifications. The CapsNet consists of primary capsule extraction, a transformation process, a routing process, and a class-conditional reconstruction, which is far more complex than ConvNets.}
        \label{fig:overview}
\end{figure*}

\paragraph{Capsule Network with Dynamic Routing~\cite{sabour2017dynamic}:} The CapsNet architecture is shown in Fig.~\ref{fig:overview}. CapsNet first extracts feature maps of shape $(C, H, W)$ from pixel intensities with two standard convolutional layers where $C$, $H$, $W$ are the number of channels, the height, and the width of the feature maps, respectively. The extracted feature maps are reformulated as primary capsules $(C/D_{in}, H, W, D_{in})$ where $D_{in}$ is the dimensions of the primary capsules. There are $M = C/D_{in} * H * W$ primary capsules in total. Each capsule $\pmb{u}_i$, a $D_{in}$-dimensional vector, consists of $D_{in}$ units across $D_{in}$ feature maps at the same location. Each primary capsule is transformed to make a vote with a transformation matrix $\pmb{W}_{ij} \in \mathbb{R}^{(D_{in} \times N*D_{out})}$, where $N$ is the number of output classes and $D_{out}$ is the dimensions of output capsules. The vote is
\begin{equation}
\pmb{\hat{u}}_{j|i} = \pmb{u}_i\pmb{W}_{ij}. 
\label{equ:vote}
\end{equation}

The routing mechanism takes all votes into consideration and identify a weight $c_{ij}$ for each vote $\pmb{\hat{u}}_{j|i}$. Concretely, the routing process iterates over the following three steps
\begin{equation}
\small
\begin{split}
\pmb{s}^{(t)}_j &= \sum^N_i c^{(t)}_{ij} \pmb{\hat{u}}_{j|i}, \\
 \pmb{v}^{(t)}_j &= g(\pmb{s}^{(t)}_j), \\
c^{(t+1)}_{ij}  &= \frac{\exp(b_{ij} + \sum_{r=1}^t \pmb{v}^{(r)}_j \pmb{\hat{u}}_{j|i}) }{\sum_k \exp(b_{ik} + \sum_{r=1}^t \pmb{v}^{(r)}_k \pmb{\hat{u}}_{k|i} )},
\end{split}
\label{eq:rounting_process}
\end{equation}
where the superscript $t$ is the index of an iteration starting from 1 and $g(\cdot)$ is a squashing function that maps the length of the vector $\pmb{s}_j$ into the range of $[0, 1)$. The $b_{ik}$ is the log prior probability. The squashing function is 
\begin{equation}
\pmb{v}_j= g(\pmb{s}_j) = \frac{\norm{\pmb{s}_j}^2}{1+\norm{\pmb{s}_j}^2} \frac{\pmb{s}_j}{\norm{\pmb{s}_j}}.
\label{equ:squa}
\end{equation}
The length of the final output capsule $\pmb{v}_j$ corresponds to the output probability of the $j$-th class. The margin loss function is applied to compute the classification loss
\begin{equation}
\begin{split}
L_k = & T_k  \max(0, m^+ - \norm{\mathbf{v}_k})^2 \\
& + \lambda (1 -T_k) \max(0, \norm{\mathbf{v}_k} -  m^-)^2
\end{split}
\label{equ:marginloss}
\end{equation}
where $T_k = 1$ if the object of the $k$-th class is present in the input. As in~\cite{sabour2017dynamic}, the hyper-parameters are often empirically set as $m^+ = 0.9$, $m^- = 0.1$ and $\lambda=0.5$.

A reconstruction sub-network reconstructs the input image from all $N$ output capsules with a masking mechanism. The ones corresponding to the non-ground-truth classes are masked with zeros before being transfered to the reconstruction sub-network. Due to the masking mechanism, only the capsule of the ground-truth class is visible for the reconstruction. Hence, the reconstruction process is called class-conditional reconstruction. The reconstruction loss is computed as a regularization term in the loss function.

\paragraph{Capsule Network Follow-Ups:} Many routing mechanisms have been proposed to improve the performance of CapsNet, such as Expectation-Maximization Routing~\cite{hinton2018matrix}, Self-Routing~\cite{hahn2019self}, Variational Bayes Routing~\cite{ribeiro2020capsule}, Straight-Through Attentive Routing~\cite{ahmed2019star}, and Inverted Dot-Product Attention routing~\cite{Tsai2020Capsules}. To reduce the parameters of CapsNet, a matrix or a tensor has been used to represent an entity instead of a vector~\cite{hinton2018matrix,rajasegaran2019deepcaps}. The size of the learnable transformation matrix can be reduced by the matrix/tensor representations. Another way to improve CapsNets is to integrate advanced modules of ConvNets into CapsNets, e.g., by skip connections~\cite{he2016deep,rajasegaran2019deepcaps} and dense connections~\cite{huang2017densely,phaye2018multi}.

Besides, the robustness of CapsNet has also been intensively investigated. Both new routing mechanisms~\cite{hinton2018matrix} and new architectures~\cite{kosiorek2019stacked} can improve the affine transformation robustness. The work~\cite{gu2020improving} achieves the best performance on the transformation robustness benchmark by simply removing the dynamic routing and by sharing the transformation matrix. The work also revealed that the high transformation robustness of CapsNets could not be attributed to the dynamic routing mechanism. The work~\cite{gu2020interpretable} replaces the dynamic routing with a multi-head attention-based graph pooling approach to achieve better interpretability. The replacement of the routing does not harm the robustness of CapsNet, even though it is the fundamental part of CapsNets. These claims further motivate us to investigate the individual components of CapsNet.

Additionally, CapsNet with new routing mechanisms can achieve high adversarial robustness~\cite{hahn2019self}. However, the work~\cite{michels2019vulnerability} shows CapsNet can be fooled as easily as CovNet. Recent work shows that the class-conditional reconstruction sub-network of CapsNet is useful to detect adversarial examples~\cite{qin2019detecting,qin2020deflecting}. The work~\cite{gu2021effective} designs the first attack method specific for CapsNet, which reduces the robust accuracy and increases the rate to pass the adversarial detection. Due to the attack-defense arms race, it is difficult to draw a solid conclusion on the adversarial robustness of CapsNet. Hence, in this work, we mainly focus on the advantage of CapsNet demonstrated in~\cite{sabour2017dynamic}.

\begin{figure*}
    \centering
    \begin{subfigure}[b]{0.31\textwidth}
        \includegraphics[width=\textwidth]{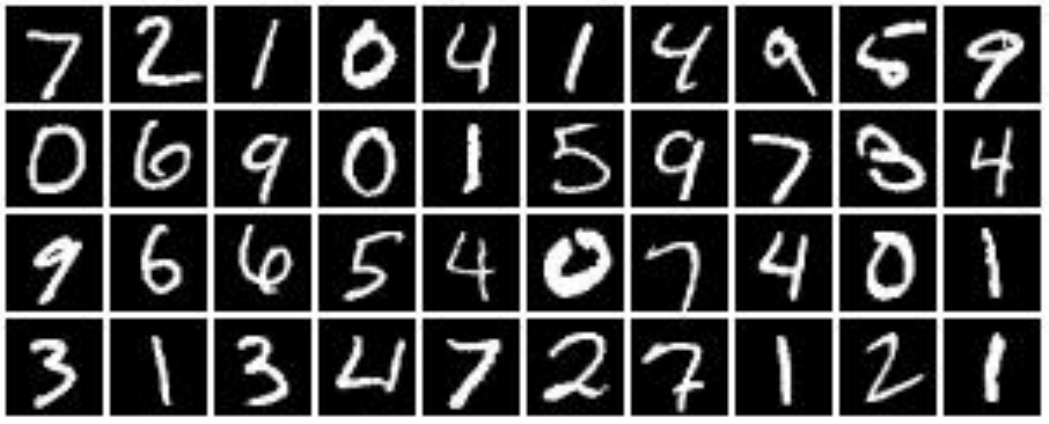}
        \caption{MNIST dataset}
        \label{fig:data_vis_mnist}
    \end{subfigure} \hspace{0.1cm}
     \begin{subfigure}[b]{0.31\textwidth}
        \includegraphics[width=\textwidth]{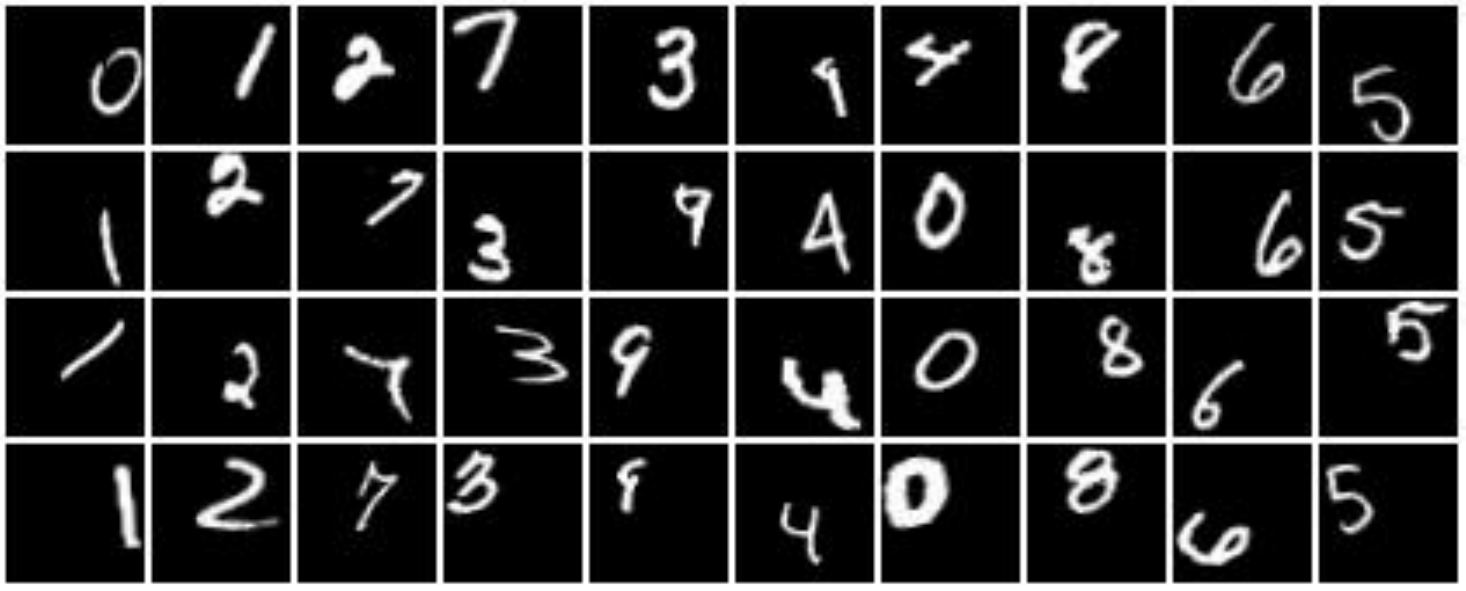}
        \caption{AffNIST dataset}
        \label{fig:data_vis_affNIST}
    \end{subfigure} \hspace{0.1cm}
     \begin{subfigure}[b]{0.31\textwidth}
        \includegraphics[width=\textwidth]{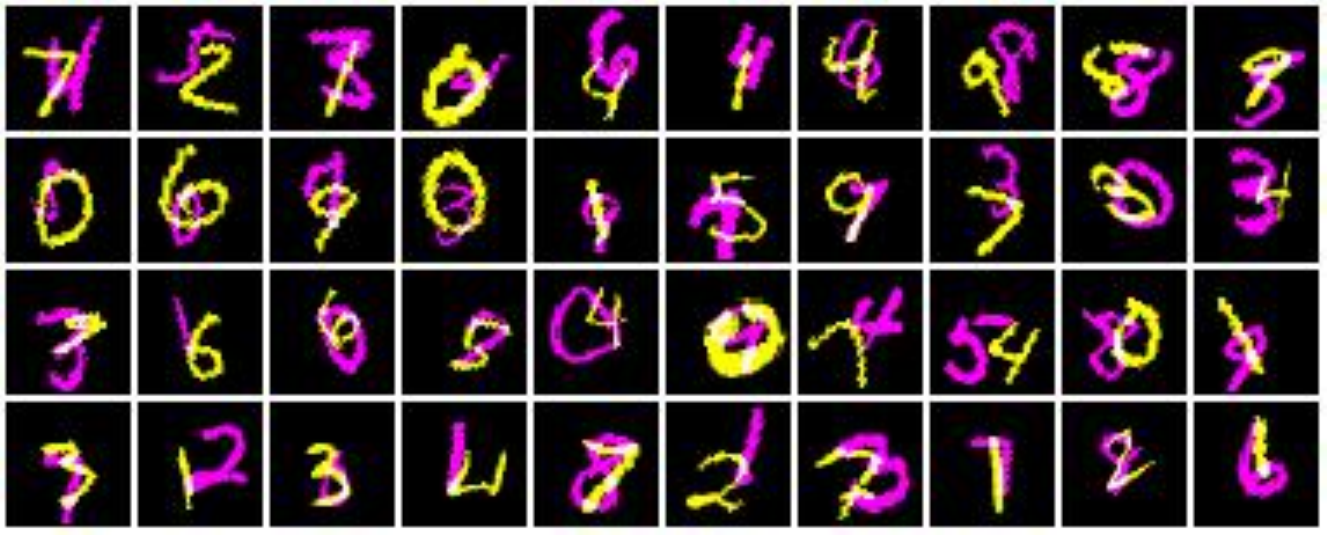}
        \caption{MultiMNIST dataset}
        \label{fig:data_vis_MultiMNIST}
    \end{subfigure}
    \caption{Visualization of datasets: While MNIST dataset corresponds to standard hand-written digits, AffNIST dataset consists of affine-transformed MNIST images. MultiMNIST dataset consists of images with two overlapping digits. In the figure, the two overlapping digits are marked with two different colors, i.e., yellow and magenta.}
    \label{fig:data_vis}
\end{figure*}

\section {Empirical Studies on Capsule Network}
\label{sec:study}

In this section, we conduct empirical studies on the robustness of CapsNets. Before we dive into the studies, we first introduce the architectures of CapsNets and ConvNets. The CapsNet we focus on in this work is Capsule Networks with dynamic routing~\cite{sabour2017dynamic}. Since the research on CapsNets is still at a primary stage, the work~\cite{sabour2017dynamic} compares their CapsNet with a LeNet-type ConvNet~\cite{lecun1998gradient}, called ConvNet-FC. The ConvNet-FC and CapsNet are illustrated in Fig.~\ref{fig:overview} on 28$\times$28 MNIST images. The notation Conv(C, K, S) stands for a convolutional layer where C, K, S are the number of channels, the kernel size, and the stride size, respectively. FC(N) is a fully connected layer where $N$ is the number of output units. All Conv and FC are followed by a ReLU activation function.

\textbf{ConvNet-FC}: The simple ConvNet baseline used in~\cite{sabour2017dynamic} is Conv(256, 5, 1) + Conv(256, 5, 1) + Conv(128, 5, 1) + FC(328) + FC(192) + Softmax(10). The three standard convolutional layers and two fully connected layers are applied to extract features from input images. An $N$-way Softmax is applied to obtain the output distribution. During training, cross-entropy loss is typically applied.

\textbf{CapsNet}: The CapsNet with Dynamic Routing in~\cite{sabour2017dynamic} is Conv(256, 9, 1) + Conv(256, 9, 2) + Dynamic Routing, followed by a reconstruction sub-network, FC(512) + FC(1024) + FC(28$\times$28). The feature maps are computed with the two standard convolutional layers. The extracted feature maps (256, H, W) is reshaped into primary capsules (32*H*W, 8) where H and W are the height and width of feature maps. The primary capsules are squashed by the squashing function in Equation (\ref{equ:squa}) and then transformed to make votes with the learned transformation matrices (32*H*W, 8, 160). The vote of each primary capsule is 160-dimensional. The dynamic routing in Equation (\ref{eq:rounting_process}) is applied to the votes to identify their weights. The output of the dynamic routing is 160-dimensional, i.e. representing 10 16-dimensional output capsules. The squashing function in Equation (\ref{equ:squa}) is applied to output capsules to map their lengths into [0, 1). The length of an output capsule is interpreted as a class output probability. In the training process, the margin loss in Equation (\ref{equ:marginloss}) is applied as the classification loss. In the class-conditional reconstruction process, a masking mechanism is applied to output capsules, where the capsules, corresponding to non-ground-truth classes, are masked with zeros. The input image is reconstructed from the masked output capsules. The reconstruction loss is used to regularize the training process.

By comparing the two networks, we can summarize 5 major differences between ConvNets and CapsNets, namely,  a transformation process, a dynamic routing layer, a squashing function, the use of a marginal loss instead of a cross-entropy loss, and a class-conditional reconstruction regularization. With these differences, CapsNet outperforms ConvNet-FC in terms of robustness to affine transformation and overlapping digits recognition as well as in learning compact semantic representations. In this section, we will investigate these advantages one by one. In each of our studies, we attempt to answer the following questions:
\vspace{-0.1cm}
\begin{enumerate}
\setlength\itemsep{0.em}
\item Do ConvNet-FC and CapsNets perform differently?
\item Which components of CapsNets make the difference?
\item How bridge the gap between the two networks?
\end{enumerate}

\subsection{Robustness to Input Affine Transformation}
\label{sec:abl_aff}

\textbf{Settings}: To examine the transformation robustness of both models, we use the popular benchmark~\cite{sabour2017dynamic,gu2020improving} where models are trained on MNIST and tested on AffNIST. In AffNIST~\cite{sabour2017dynamic}, the original 28$\times$28 MNIST images are first padded with 6 pixels to 40$\times$40 image and then affine transformed, namely, rotation within 20 degrees, shearing within 45 degrees, scaling from 0.8 to 1.2 in both vertical and horizontal directions, and translation within 8 pixels in each direction. In the training dataset, the 28$\times$28 MNIST images are placed randomly on a black background of 40$\times$40 pixels without further transformation. The image examples are visualized in Fig.~\ref{fig:data_vis}. The performance on both MNIST and AffNIST test datasets is reported. All scores are averaged over 5 runs across this paper.

Besides ConvNet-FC and CapsNet, we include the state-of-the-art model on the benchmark in this experiment, namely, Aff-CapsNet. It simplifies CapsNet by removing dynamic routing and sharing the transformation matrix in the transformation process. 

Following~\cite{sabour2017dynamic,gu2020improving}, the Adam optimizer is used to train the models with an initial learning rate of 0.001 and a batch size of 128. In CapsNet, the reconstruction loss is scaled down by 0.0005 so that it does not dominate the margin loss during training. It is hard to decide which model is more robust to affine transformations when they achieved different accuracy on untransformed examples. To eliminate this confounding factor, we stopped training the models when they achieve similar performance (i.e. about 99.22\%), following~\cite{sabour2017dynamic}.

\begin{table}[h]
\begin{center}
\footnotesize
\setlength\tabcolsep{0.14cm}
\begin{tabular}{P{2.8cm}P{1.4cm}P{1.4cm}P{1.4cm}}
\toprule
Models  & \#Para. &  MNIST &  AffNIST \\
\midrule
GE-CapsNet~\cite{lenssen2018group}& -  & 98.42 & 89.10 \\
SPARSECAPS~\cite{rawlinson2018sparse} & -  & 99 & 90.12 \\
SCAE~\cite{kosiorek2019stacked} & -  & 98.5 & 92.21 \\
EM-CapsNet~\cite{hinton2018matrix} & -  & 99.2 & 93.1 \\
\midrule
ConvNet-FC~\cite{sabour2017dynamic} & 35.4M  & 99.22  & 66  \\
CapsNet~\cite{sabour2017dynamic} & 13.5M  & 99.23 &  79  \\
CapsNet-NoR~\cite{gu2020improving} & 13.5M  & 99.22 &  81.81  \\
Aff-CapsNet-DR~\cite{gu2020improving}  & 7.5M  & 99.22 & 89.03  \\
Aff-CapsNet~\cite{gu2020improving} & 7.5M  & 99.23 & 93.21 \\
\midrule
\textbf{ConvNet-Avg} & 5.3M & 99.22 & \textbf{94.11}  \\
\bottomrule
\end{tabular}
\end{center}
\caption{Comparison on the transformation robustness benchmark: The generalization performance to AffNIST is reported when models achieve similar performance on MNIST test dataset. Our simple ConvNet-Avg is more robust than CapsNet to input affine transformations.}
\label{tab:affNoRSOTA}
\vspace{-0.1cm}
\end{table}

\begin{table*}[t]
\begin{center}
\footnotesize
\begin{tabular}{c | c | c | c | c | c | c | c | c}
\hline
Factors & Routing & Shared TransM & Squash-fn & Reconstion & Loss & Train-MNIST & Test-MNIST & Test-AffNIST \\
\hline
\hline
\multirow{2}{*}{Routing} & \textcolor{colormark}{NoR} & \checkmark  & \checkmark  &  \checkmark & MarginLoss & 100 & 99.29{\tiny ($\pm$ 0.13)}  & 93.55{\tiny ($\pm$ 1.47)} \\
\cline{2-9}
 &  \textcolor{colormark}{DR} & - & -  & - & -   & 100 & 99.21{\tiny ($\pm$ 0.31)} & \textbf{90.07}{\tiny ($\pm$ 0.98)} \\
\hline
\hline
\multirow{2}{*}{Shared TransM} & NoR & \textcolor{colormark}{$\pmb{\checkmark}$}  & \checkmark  &  \checkmark & MarginLoss & 100 & 99.29{\tiny ($\pm$ 0.13)}  & 93.55{\tiny ($\pm$ 1.47)} \\
\cline{2-9}
&  - & \textcolor{colormark}{$\pmb{\times}$} & -  & - & -  & 100 & 98.98{\tiny ($\pm$ 0.04)} & \textbf{80.49}{\tiny ($\pm$ 0.34)} \\
\hline
\hline
\multirow{2}{*}{Squash-fn}  & NoR & \checkmark  & \textcolor{colormark}{$\pmb{\checkmark}$}  &  \checkmark & MarginLoss & 100 & 99.29{\tiny ($\pm$ 0.13)}  & 93.55{\tiny ($\pm$ 1.47)} \\
\cline{2-9}
 &  - & - & \textcolor{colormark}{$\pmb{\times}$} & - & -  & 99.75 & 97.93{\tiny ($\pm$ 0.13)} & \textbf{80.42}{\tiny ($\pm$ 0.39)} \\
\hline
\hline
\multirow{3}{*}{Reconstruction} &  NoR & \checkmark  & \checkmark  & \textcolor{colormark}{conditional} & MarginLoss  & 100 & 99.29{\tiny ($\pm$ 0.13)}  & 93.55{\tiny ($\pm$ 1.47)} \\
\cline{2-9}
 & - & - & -  & \textcolor{colormark}{normal} & -  & 100 & 99.43{\tiny ($\pm$ 0.28)} & 95.09{\tiny ($\pm$ 0.56)} \\
\cline{2-9}
 & - & - & -  & \textcolor{colormark}{$\pmb{\times}$} & -  & 100 & 99.39{\tiny ($\pm$ 0.26)} & 93.49{\tiny ($\pm$ 0.46)} \\
\hline
\hline
\multirow{2}{*}{Loss} & NoR & \checkmark  & \checkmark  &  \checkmark & \textcolor{colormark}{MarginLoss} & 100 & 99.29{\tiny ($\pm$ 0.13)}  & 93.55{\tiny ($\pm$ 1.47)} \\
\cline{2-9}
 & - & - & -  & - & \textcolor{colormark}{CE Loss} & 100 & 99.27{\tiny ($\pm$ 0.05)} & 94.67{\tiny ($\pm$ 0.43)} \\
\hline
\end{tabular}
\end{center}
\caption{The performance on MNIST training dataset, MNIST test dataset, and AffMNIST test dataset are reported, respectively (in percentage \%). Dynamic Routing (DR) and Margin loss are even harmful to the transformation robustness, while the squashing function (Squash-fn) and the shared transformation matrix (Shared TransM) are beneficial.}
\label{tab:perf_affcaps}
\end{table*}

\textbf{Results and Analysis}: The performance is reported in Tab.~\ref{tab:affNoRSOTA}. We can observe that there is a gap between ConvNet-FC and CapsNet. As reported in~\cite{sabour2017dynamic,gu2020improving}, the CapsNet outperforms ConvNet-FC, and Aff-CapsNet outperforms CapsNet. We take Aff-CapsNet (a simplified CapsNet) as a baseline and conduct further ablation studies on the components of CapsNet in Tab.~\ref{tab:perf_affcaps}. We report the model test performance on both un-transformed MNIST test images and novel affine-transformed ones. No early stopping is applied in the ablation studies.

The transformation process can be seen as a fully connected (FC) layer since the transformation matrices therein are equivalent to the parameters of an FC layer. \textit{Why is Aff-CapsNet more robust than CapsNet?} The transformation robustness of CapsNet can be improved by sharing the transformation matrix. When the transformation matrix is shared and no routing is applied in Aff-CapsNet, the transformation process is essential to conduct group $1\times1$ convolutional operations, global average pooling operations, and an average operation on the pooling results of different groups. A further study shows that the number of groups has no effect on the robustness (see Supplement A). Hence, we attribute the superior performance of the sharing transformation matrix to the global average pooling operation. \textit{Why is CapsNet more robust than ConvNet-FC?} The ConvNet-FC has two fully connected layers, while CapsNet has a functionally similar one. Another difference between them is the kernel size. Our study shows that large kernels are also beneficial to achieve transformation robustness (see Tab.~\ref{tab:perf_affcaps_kernel}). This argument also echoes our claim above. Namely, both global average pooling and large kernels improve the robustness by increasing receptive fields.

In Tab.~\ref{tab:perf_affcaps}, the dynamic routing is even harmful to the transformation robustness, which is also supported by the Tab.~\ref{tab:affNoRSOTA}. In addition, when no squashing function is applied, CapsNet has to regress the capsule length to extreme values (e.g., 0 or 1), which is a hard task and leads to unsatisfying performance (even on the training dataset). The margin loss can slightly weaken the transformation robustness of CapsNet, while reconstruction makes no difference to it. The non-conditional reconstruction slightly improves the performance since it updates all capsules in each training iteration.

Based on our findings, we propose a new simple ConvNet baseline, called \textbf{ConvNet-Avg}. It starts with the two convolutional layers and terminates with a global average pooling and an output layer, which is also a common architecture used in image classification. The cross-entropy loss is applied to train the model. To make a fair comparison, we use the same convolutional layers as in CapsNet and Aff-CapsNet, namely, Conv(256, 9, 1) + Conv(256, 9, 2) + Global AvgPool + FC(10) (see Fig.~\ref{fig:overview}). It is hard to decide which model is better at generalizing to affine transformations when they achieved different accuracy on untransformed examples. We follow previous work and stop training the models when they achieve similar test performance (99.22\%). As shown in Tab.~\ref{tab:affNoRSOTA}, our simple ConvNet-Avg achieves slightly better performance with fewer parameters.

\textbf{Conclusions}: 1) Compared to ConvNet-FC, CapsNet achieves better test performance with fewer parameters on AffNIST. We attribute the gap to the kernel size. 2) Dynamic routing can harm the transformation robustness of CapsNet. When the routing is removed, the uniform average of votes (i.e., NoR) aggregates the global information better. 3) Our baseline ConvNet-Avg outperforms CapsNets significantly. It consists of only convolutional layers and a global average pooling layer, and no advanced component from SOTA ConvNets. The simplicity of ConvNet-Avg indicates that CapsNets are even less robust to affine transformation than ConvNets in a fair comparison.

\begin{table*}[t]
\begin{center}
\footnotesize
\begin{tabular}{c | P{0.6cm}P{0.6cm}P{0.6cm} | P{0.6cm}P{0.6cm}P{0.6cm} | P{0.6cm}P{0.6cm}P{0.6cm} | P{0.6cm}P{0.6cm}P{0.6cm} | P{0.6cm}P{0.6cm}P{0.6cm}}
\hline
Kernels  & \multicolumn{3}{c|}{\textcolor{colormark}{K(3, 3)}} &  \multicolumn{3}{c|}{\textcolor{colormark}{K(5, 5)}}  &  \multicolumn{3}{c|}{\textcolor{colormark}{K(7, 7)}} & \multicolumn{3}{c|}{\textcolor{colormark}{K(9, 9)}} &  \multicolumn{3}{c}{\textcolor{colormark}{K(11, 11)}} \\
\hline
Models  & \#Para. & A$_{std}$ & A$_{aff}$ & \#Para. & A$_{std}$ & A$_{aff}$ & \#Para. & A$_{std}$ & A$_{aff}$ & \#Para. & A$_{std}$ & A$_{aff}$ & \#Para. & A$_{std}$ & A$_{aff}$  \\
\hline
CapsNet & 16.1M  & 96.31 & 61.36  & 14.4M  & 98.18 & 70.34  & 13.5M  & 98.74 & 75.82  &13.5M  & 99.26 & 79.12  & 14.3M  & 99.1 & 86.79  \\
\hline
ConvNet-FC & 49.5M  & 96.54 & 64.57  & 35.4M  & 99.23 & 66.08  & 25.2M  & 99.03 & 66.76  & 18.8M  & - & -  & 16.19M  & - & -  \\
\hline
ConvNet-Avg & 0.59M  & 97.14 & \textbf{86.58}  & 1.70M  & 98.58 & \textbf{90.95}  & 3.23M  & 99.1 & \textbf{92.31}  & 5.30M  & 99.22 & \textbf{94.11} & 7.96M  & 99.34 & \textbf{90.58}  \\
\hline
\end{tabular}
\end{center}
\caption{The effect of the kernel sizes on the transformation robustness of different models: Both standard accuracy (A$_{std}$) and the generalization accuracy (A$_{aff}$) on transformed data are reported. The large kernels make positive contributions to the transformation robustness. When the same kernel size is applied, ConvNet-Avg outperforms both ConvNet-FC and CapsNet.}
\label{tab:perf_affcaps_kernel}
\end{table*}

\subsection{Recognizing overlappping digits}
\label{sec:over_digits}

\textbf{Settings}: The work~\cite{sabour2017dynamic} shows that the CapsNet is able to recognize overlapping digits by segmenting them. To check this property, we use the MultiMNIST dataset, which is generated by overlaying a digit on top of another digit but from a different class. Specifically, a 28$\times$28 MNIST image with a digit is first shifted up to 4 pixels in each direction resulting in a 36$\times$36 image. The resulting image is overlaid to another image from different classes but the same set (training dataset or test dataset). For each image in MNIST, we can create $N$ (from 1 to 1K) images. See Fig.~\ref{fig:data_vis_MultiMNIST} for some examples from data.

The classification of an image with overlapping digits is correct if both digits are correctly classified (the top 2 output classes match the ground truth). The margin loss can be applied to compute the classification loss. In the ConvNet baselines, the sigmoid function is applied to logits instead of softmax to obtain output probabilities since this is a multi-target classification task, and the binary cross-entropy loss is applied to compute the classification loss.

In the training process, the CapsNet is first applied to the overlapping digits to obtain output capsules.  During reconstruction, a ground-truth class is picked at a time, and the capsule corresponding to the class is kept for the reconstruction while others are masked with zeros. In other words, we run the reconstruction sub-network twice, each for one digit. The reconstruction loss can be computed similarly since the images of individual digits are available.

\begin{table*}[t]
\begin{center}
\footnotesize
\begin{tabular}{c | c | c | c | c | c | c | c }
\hline
Factors & Routing & Shared TransM & Squash-fn & Reconstion & Loss & Train-MultiMNIST & Test-MultiMNIST \\
\hline
\hline
\multirow{2}{*}{Routing} & \textcolor{colormark}{DR} & $\pmb{\times}$  & \checkmark  &  \checkmark & MarginLoss & 94.03 & 93.26{\tiny ($\pm$ 0.24)} \\
\cline{2-8}
 &  \textcolor{colormark}{NoR} & - & -  & - & -   & 90.28 & 90.07{\tiny ($\pm$ 0.29)} \\
\hline
\hline
\multirow{2}{*}{Shared TransM} & DR & \textcolor{colormark}{$\pmb{\times}$}  & \checkmark  &  \checkmark & MarginLoss & 94.03 & 93.26{\tiny ($\pm$ 0.24)} \\
\cline{2-8}
&  - & \textcolor{colormark}{$\pmb{\checkmark}$} & -  & - & -  & 86.92 & \textbf{86.44}{\tiny ($\pm$ 0.37)} \\
\hline
\hline
\multirow{2}{*}{Squash-fn}  & DR & $\pmb{\times}$  & \textcolor{colormark}{$\pmb{\checkmark}$}  &  \checkmark & MarginLoss & 94.03 & 93.26{\tiny ($\pm$ 0.24)} \\
\cline{2-8}
 &  - & - & \textcolor{colormark}{$\pmb{\times}$} & - & -  & 87.71 & \textbf{87.24}{\tiny ($\pm$ 0.53)} \\
\hline
\hline
\multirow{3}{*}{Reconstruction} &  DR & $\pmb{\times}$  & \checkmark  & \textcolor{colormark}{conditional} & MarginLoss  & 94.03 & 93.26{\tiny ($\pm$ 0.24)} \\
\cline{2-8}
 & - & - & -  & \textcolor{colormark}{normal} & -  & 93.83 & 93.19{\tiny ($\pm$ 0.30)}  \\
 \cline{2-8}
 & - & - & -  & \textcolor{colormark}{$\pmb{\times}$} & -  & 90.28 & 90.17{\tiny ($\pm$ 0.26)}  \\
\hline
\hline
\multirow{2}{*}{Loss} & DR & $\pmb{\times}$  & \checkmark  &  \checkmark & \textcolor{colormark}{MarginLoss} & 94.03 & 93.26{\tiny ($\pm$ 0.24)} \\
\cline{2-8}
 & - & - & -  & - & \textcolor{colormark}{BCE Loss} & 91.64 & 91.19{\tiny ($\pm$ 0.35)} \\
\hline
\end{tabular}
\end{center}
\caption{The ablation study on components of CapsNet: The performance of models trained on 6M overlapping digits. All individual components make positive contributions to the ability to recognize overlapping digits. The transformation matrices contribute the most; the performance drops dramatically if a shared transformation matrix is applied.}
\label{tab:perf_multi_mnist}
\end{table*}

\textbf{Results and Analysis}: The overlapping digit recognition performance is reported in Tab.~\ref{tab:perf_multi_mnist} where the individual components of CapsNets are ablated. The reconstruction sub-network helps to improve the recognition performance. However, it does not have to be class-conditional. The reconstruction loss regularizes the training process so that the information about both digits is encoded in features and high-level capsules. The margin loss can be directly applied to a multi-target classification task, which outperforms the standard binary cross-entropy loss. Both the reconstruction and the margin loss can be applied to enhance a ConvNet.

When a vector representation is applied, the squashing function plays an important role. When applying the squashing function to the primary capsules, the feature maps are group-wise normalized. The information is communicated across different channels, which can help to better disentangle overlapping digits. Additionally, CapsNet has to regress the non-squashed capsule length to certain values. Since the regression task is hard, CapsNets achieve unsatisfying performance on both the training and test dataset. The analysis echoes the one in Sec.~\ref{sec:abl_aff}.

The dynamic routing process identifies the weights for votes, which results in a higher modeling capacity than the uniform averaging operation on votes. Other components that support the CapsNet's modeling capacity are the transformation matrices. When a shared transformation matrix is applied, the model performance drops dramatically. We check the ConvNet-Avg on this task and observe that CapsNet outperforms ConvNet-Avg significantly. The reason behind this is that the global pooling operation can be harmful for recognizing overlapping digits since it aggregates a feature map into a single unit. The convolutional layer itself is not able to disentangle the overlapping digits into different feature maps. In CapsNet, the transformation process acts as a fully connected layer, which avoids the global average pooling. Hence, we argue that the high modeling capacity is the essential reason why CapsNet performs well on the overlapping digits recognition task.

\begin{figure}[h]
    \centering
        \includegraphics[width=0.4\textwidth]{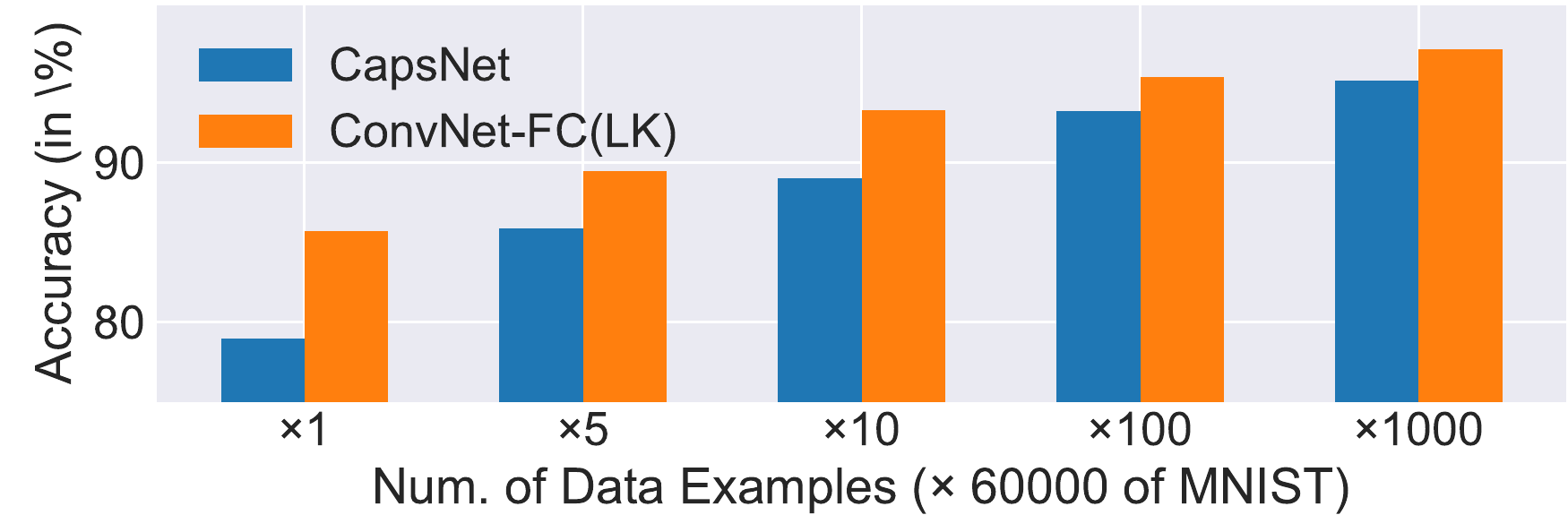}
        \caption{ConvNet-FC(LK) outperforms CapsNet on MultiMNIST dataset with different data sizes.}
         \label{fig:datasize_MultiMNIST}
\end{figure}

FC layers in ConvNet-FC can maintain richer information (features at all locations) for distinguishing overlapping digits. Note that the baseline ConvNet-FC in~\cite{sabour2017dynamic} has a smaller kernel size than in CapsNet. Hence, we propose to apply ConvNet-FC with large kernels (ConvNet-FC(LK)) to this overlapping digits recognition task. In ConvNet-FC(LK), we also reduce the units of fully connected layers to save parameters so that it can be compared to CapsNets. When the same large kernel is applied, ConvNet-FC(LK) outperforms the CapsNet and sets a new SOTA on this benchmark (97.11\% vs. 95.18\%). When different training data sizes and different kernel sizes are applied in the experiments, the simple ConvNet-FC(LK) outperforms the CapsNet consistently (See Fig.~\ref{fig:datasize_MultiMNIST} and Supplement B).

\textbf{Conclusions}: 1) All the components contribute to the ability of CapsNet to recognize overlapping digits. 2) The transformation process with a non-shared transformation matrix and a dynamic routing to weight votes bring high modeling capacity, which essentially supports the high performance of CapsNet in this task. 3) The simple ConvNet-FC(LK) with similar parameters performs better than CapsNet on this benchmark, which indicates that CapsNet is not more robust than ConvNet to recognize overlapping digits.

\subsection{Semantic Capsule Representations}
\textbf{Settings:} In CapsNets, when a single element in a capsule is perturbated, the reconstructed images are visually changed correspondingly~\cite{sabour2017dynamic}, see Fig.~\ref{fig:ConvNet_Caps_perb}. The visual changes often correspond to human-understandable semantic object variations. In this experiment, we investigate which components support the semantic representations. Since this property is mainly demonstrated by a reconstruction sub-network, we introduce three models below:

\begin{figure*}
    \centering
    \begin{subfigure}[b]{0.23\textwidth}
        \includegraphics[width=\textwidth]{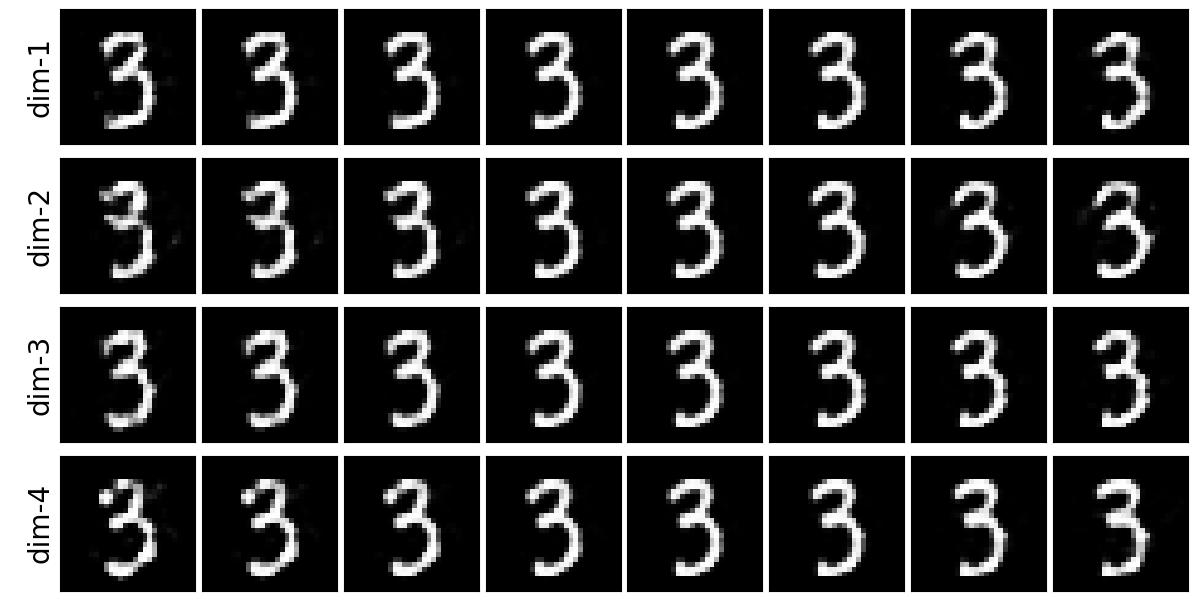}
        \caption{on ConvNet-R}
        \label{fig:ConvNet_R_perb}
    \end{subfigure}\hspace{0.2cm}
     \begin{subfigure}[b]{0.23\textwidth}
        \includegraphics[width=\textwidth]{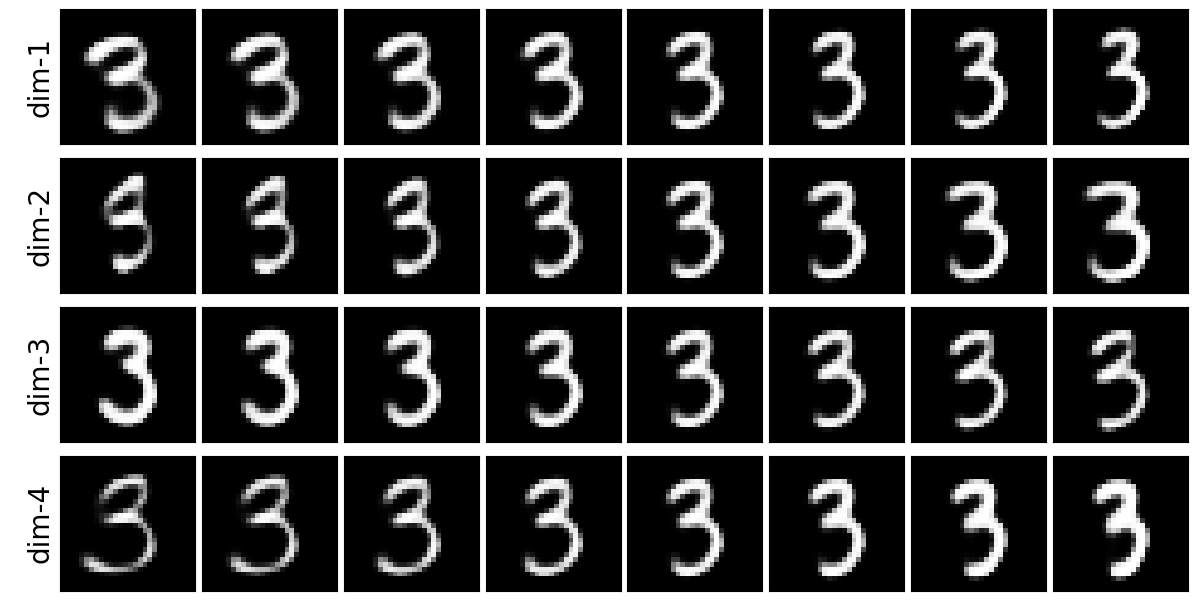}
        \caption{on ConvNet-CR}
        \label{fig:ConvNet_CR_perb}
    \end{subfigure}\hspace{0.2cm}
     \begin{subfigure}[b]{0.23\textwidth}
        \includegraphics[width=\textwidth]{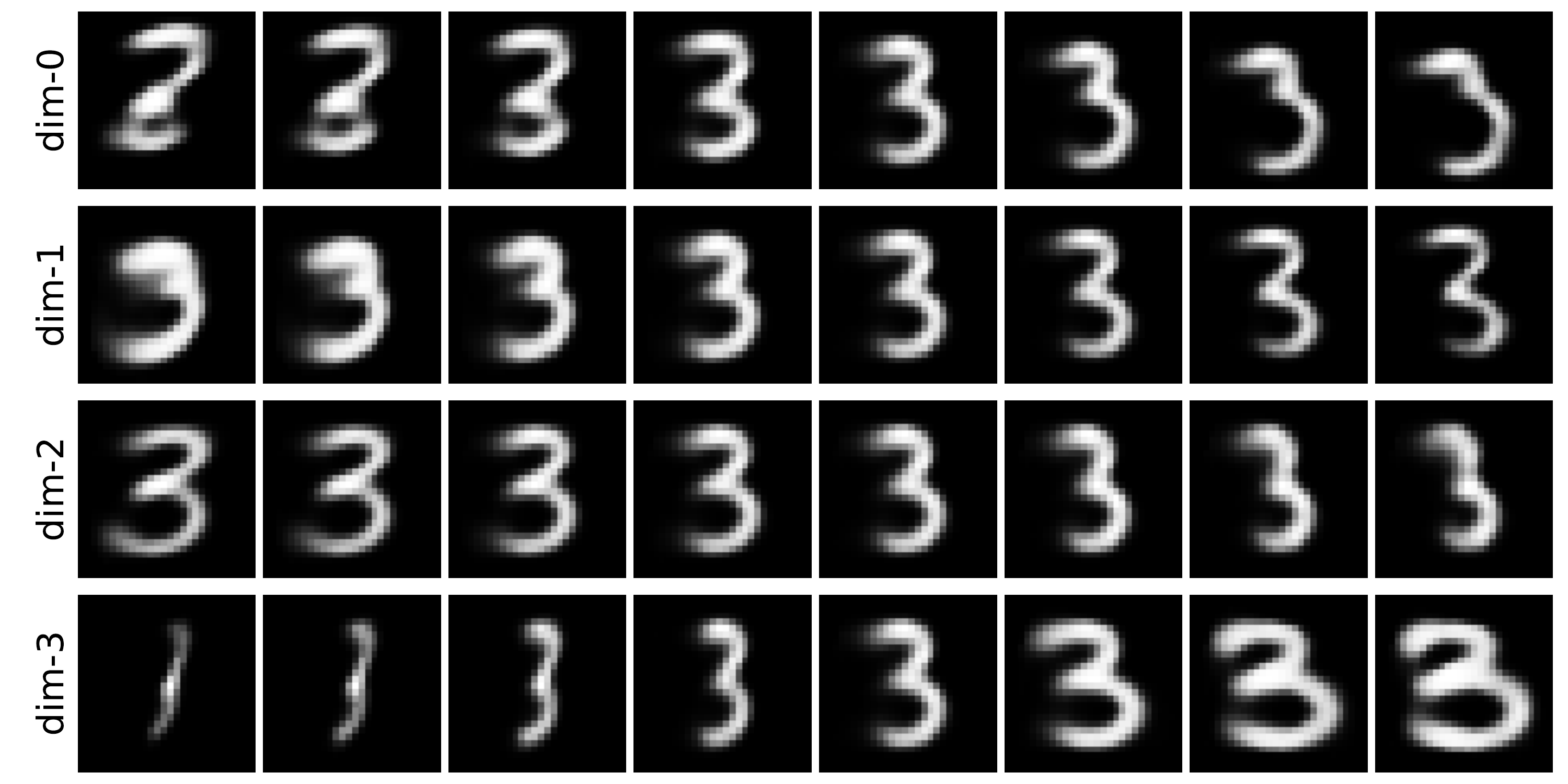}
        \caption{on ConvNet-CR-SF}
        \label{fig:ConvNet_CR_SF_perb}
    \end{subfigure}\hspace{0.2cm}
     \begin{subfigure}[b]{0.23\textwidth}
        \includegraphics[width=\textwidth]{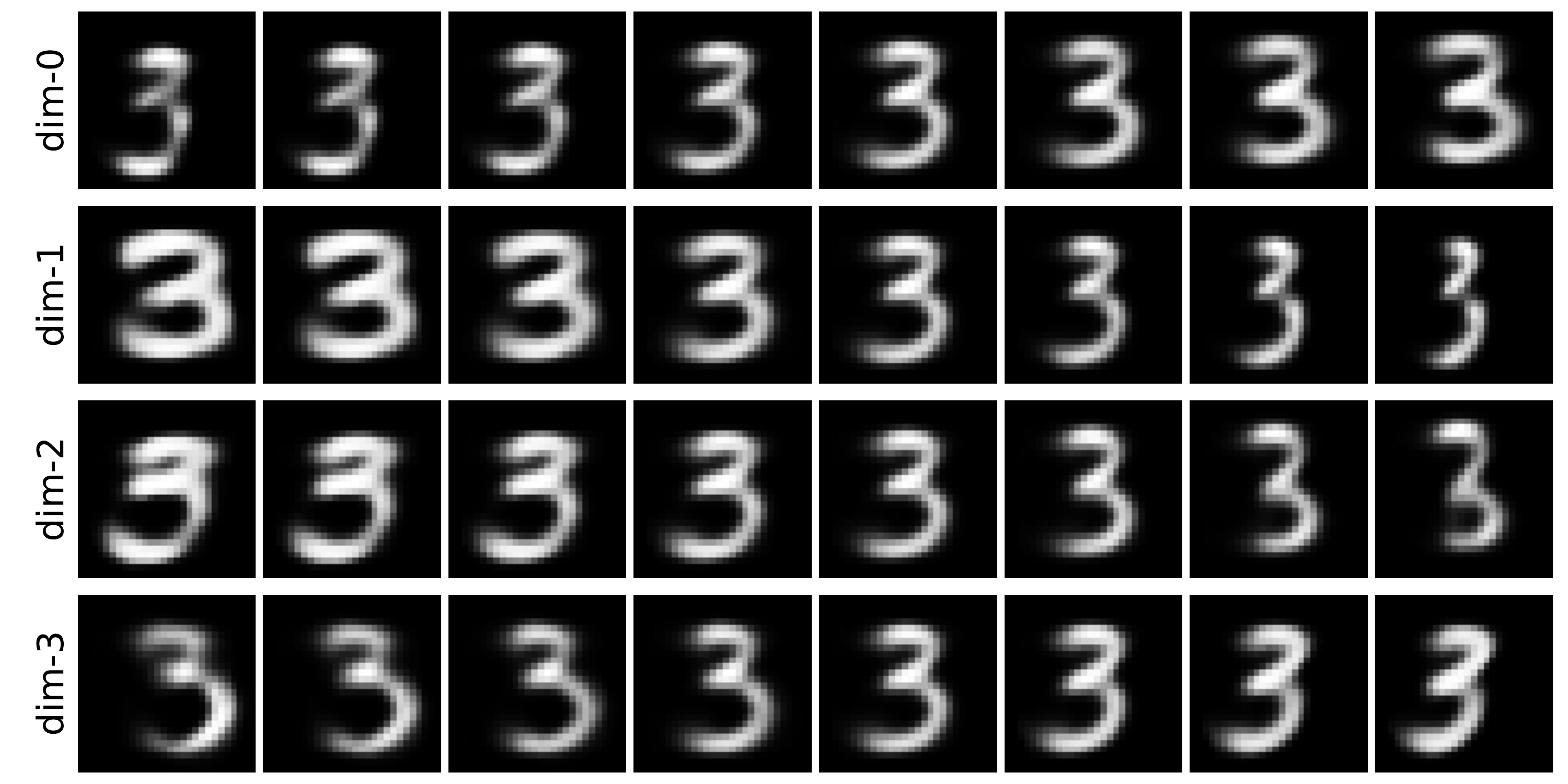}
        \caption{on CapsNet}
        \label{fig:ConvNet_Caps_perb}
    \end{subfigure}
    \caption{The reconstructed images are shown when a single unit is perturbed. The reconstruction only helps when the class-conditional masking mechanism is applied. The squashing function improves the visual response further.}
    \label{fig:comapct_rep_vis}
\end{figure*}

\begin{figure*}
    \centering
    \begin{subfigure}[b]{0.32\textwidth}
        \includegraphics[width=\textwidth]{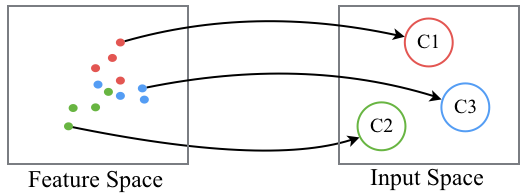}
        \caption{Reconstruction in ConvNet-R}
        \label{fig:ConvNet_R_recons}
    \end{subfigure} \hspace{0.1cm}
     \begin{subfigure}[b]{0.32\textwidth}
        \includegraphics[width=\textwidth]{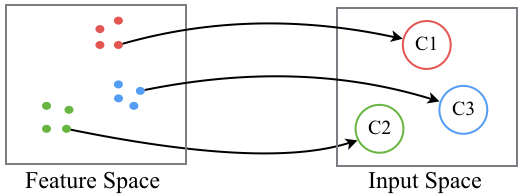}
        \caption{Reconstruction in ConvNet-CR}
        \label{fig:ConvNet_CR_recons}
    \end{subfigure} \hspace{0.1cm}
     \begin{subfigure}[b]{0.32\textwidth}
        \includegraphics[width=\textwidth]{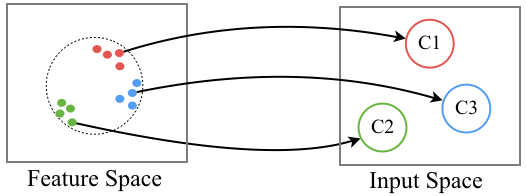}
        \caption{Reconstruction in ConvNet-CR-SF}
        \label{fig:ConvNet_CR_SF_recons}
    \end{subfigure}
    \caption{The reconstruction from feature space to the input space: In ConvNet-R, the capsule representations of different classes are entangled in feature space; the ones in ConvNet-CR are clearly separated due to the class-conditional masking mechanism. When a squashing function is applied to squash the vector, the representations live within a manifold. The representation constraints improve the network's ability to extrapolate object variations.}
    \label{fig:recon_analysis}
\end{figure*}

\textbf{ConvNet-CR}: This ConvNet baseline has the same number of parameters as in CapsNet and the same reconstruction sub-network. Its architecture is Conv(256, 9, 1) + Conv(256, 9, 2) + FC(160), where 160 corresponds to the dimensions of output capsules and the parameters in FC(160) corresponds to the non-shared transformation matrices of CapsNet. The 160 activations are grouped into 10 groups where each group corresponds to an output capsule. The sum of 16 activations in each vector corresponds to a logit. The sigmoid function is applied to each logit to obtain the output probability. The reconstruction sub-network reconstructs the input from (the 160 activations) with a masking mechanism, similar to that in CapsNet. 

\textbf{ConvNet-R}: In this baseline, an output layer FC(10) is built on the 160 activations of ConvNet-CR instead of grouping them. The reconstruction sub-network of ConvNet-CR reconstructs the input from the 160 activations directly, without the masking mechanism. 

\textbf{ConvNet-CR-SF}: This baseline equips ConvNet-CR with the squashing function in Equation (\ref{equ:squa}). The feature maps from Conv(256, 9, 2) are mapped into vectors with the same shape of primary capsules, and the vectors are squashed. Each element of the vectors is fully connected to 160 units of the next layer. The 160 activations are grouped to obtain the 10 output vectors. The vectors are similarly squashed so that their lengths stand for the output probability of the corresponding class. This baseline is equivalent to CapsNet without a routing mechanism (CapsNet-NoR).

In CapsNet, several units can correspond to a similar semantic concept. An interesting question to investigate is, what percentage of neurons strongly react to changes of a given latent factor. We propose a metric to evaluate such compactness. Given a latent factor $z$ (e.g. rotation) and an image X, we compute the semantic compactness score with the following steps:
\begin{enumerate}
\setlength\itemsep{0.em}
\item \hspace{-0.15cm} Creating a list of images with different rotation degrees;
\item \hspace{-0.15cm} Obtaining their representation vectors via forward inferences (the vectors of ground-truth classes are kept);
\item \hspace{-0.15cm} Computing the variance of the vectors in each dimension $\pmb{Var}$ and normalize them by their sum $\pmb{Var}_n$;
\item \hspace{-0.15cm} Computing the KL divergence between the normalized variance values $\pmb{Var}_n$ and a uniform prior. 
\end{enumerate}
\vspace{-0.1cm}
The compactness score is averaged over the whole dataset. The higher the score is, the more compact the semantic representation becomes. The intuition behind the score is that, if only one unit changes when images are rotated, the normalized variance will be one-hot, and the relative entropy to uniform prior is the maximum.

\textbf{Results and Analysis}: After training, we perform the capsule perturbation experiments on the 160 activations, as in~\cite{sabour2017dynamic}. In CapsNet, we tweak one dimension of capsule representations by intervals of 0.05 in the range [-0.2, 0.2]. The reconstructed images are visualized in Fig.~\ref{fig:ConvNet_Caps_perb}. The semantic changes of images can be observed, e.g., the rotation and the stroke thickness. We find that the reconstructed images in ConvNets stay almost unchanged visually when perturbing the corresponding activation with the same range. The observation can be caused by the too-small perturbation range for the unit activations. Hence, we increase the range gradually until the reconstructed image cannot be recognized where we reach the range of [-8, 8]. The reconstructed images are shown in Fig.~\ref{fig:comapct_rep_vis}. In ConvNet-R, the semantics of reconstructed images is not sensitive to all individual dimensions in Fig.~\ref{fig:ConvNet_R_perb}. In ConvNet-CR, where the class-conditional reconstruction is applied, the changes of representation unit also cause the semantic changes of reconstructed images in Fig.~\ref{fig:ConvNet_CR_perb}. When the squashing function is applied, the representations in ConvNet-CR-CF strongly react to the perturbations in Fig.~\ref{fig:ConvNet_CR_SF_perb}.

\begin{table}[t]
\begin{center}
\footnotesize
\setlength\tabcolsep{0.06cm}
\begin{tabular}{P{1.9cm}P{0.cm}P{1.cm}P{1.cm}P{0.8cm}P{1.cm}P{1.cm}}
\toprule
Datasets  & \multicolumn{6}{|c}{MNIST} \\
\midrule
Factors  & \multicolumn{1}{|c}{Rotation}  & Trans-X &  Trans-Y &  Scale  & Shear-X & Shear-Y \\
\midrule
ConvNet-R               &  \multicolumn{1}{|c}{0.0003}  & 0.0016  & 0.0009 & 0.0004  & 0.0003 & 0.0007  \\
\midrule
ConvNet-CR           &   \multicolumn{1}{|c}{0.0028}  & 0.0038  & 0.0032 & 0.0052  & 0.0058 & 0.0022  \\
\midrule
ConvNet-CR-SF       &  \multicolumn{1}{|c}{\textbf{0.0325}}  & \textbf{0.2010}  & \textbf{0.3192} & \textbf{0.0146}  & \textbf{0.0476} & \textbf{0.0506} \\
\midrule
CapsNet            &  \multicolumn{1}{|c}{0.0031}  & 0.0107  & 0.0464 & 0.0026  & 0.0098 & 0.0021 \\
\bottomrule
\end{tabular}
\end{center}
\caption{The representation compactness: The class-conditional reconstruction and the squashing function improve the compactness, while dynamic routing reduces it.}
\label{tab:comapct_rep_eval}
\end{table}

Both the class-conditional reconstruction mechanism and the squashing function can help ConvNets to learn meaningful semantic representations. The two components characterize the function learned by the reconstruction sub-network, which maps representations from feature space back to input space. We illustrate the characteristics of these functions in Fig.~\ref{fig:recon_analysis}, using an example with a 2D input space and 3 output classes. The ConvNet-R reconstructs inputs from the features that are entangled to some degree. In ConvNet-CR, the features of different classes are perfectly separated since the features are class-conditional. The ConvNet-CR-CF constrains the feature space further by squashing the vectors so that they live inside a manifold. We also report the compactness score of each model in Tab.~\ref{tab:comapct_rep_eval}. We speculate that it is these constraints that improve the representation's compactness. More experiments on the FMNIST dataset can be found in Supplement C.

\textbf{Conclusions}: Both the class-conditional reconstruction and the squashing function help CapsNet learn meaningful semantic representations, while dynamic routing is even harmful. The two components can be integrated into ConvNets, where ConvNet-CR-SF learns better semantic compact representations than CapsNets.

\section{Conclusion}
We reveal 5 major differences between CapsNets and ConvNets and study 3 properties of CapsNets. We show that dynamic routing is harmful to CapsNets in terms of transformation robustness and semantic representations. In each presented task, a simple ConvNet can be built to outperform the CapsNet significantly. We find that there is no single ConvNet that can outperform CapsNet in all cases. Hence, we conclude that \textit{CapsNets with dynamic routing are not more robust than ConvNets}. We leave further explorations for future work, e.g., concerning different datasets, and other properties of CapsNets, and other CapsNets.

The dynamic routing aggregates information from low-level entities into high-level ones. The aggregation can be also be done by a graph pooling operation~\cite{gu2020interpretable}. In ConvNets, the relationship between low-level entities is also explored in aggregation~\cite{hu2018relation,hu2019local}. More aggregation approaches wil be explored in future work.

{\small
\bibliographystyle{ieee_fullname}
\bibliography{egbib}
}

\end{document}